\RequirePackage{amsmath}
\documentclass[runningheads]{llncs}
\usepackage[T1]{fontenc}
\usepackage{graphicx}
\usepackage{graphicx}
\usepackage{algorithmic}
\usepackage{amssymb}
\usepackage{algorithm}
\usepackage{color}
\usepackage{siunitx}
\usepackage{booktabs}
\usepackage{threeparttable}
\usepackage{multirow}
\usepackage{marvosym}
\usepackage{amsmath}
\newcommand{\ie}{\textit{i}.\textit{e}.}
\newcommand{\eg}{\textit{e}.\textit{g}.}

\usepackage[colorlinks,hyperindex,breaklinks]{hyperref}
\usepackage{amssymb} 
\usepackage{pifont} 
\newcommand{\cmark}{\ding{51}}

\newcommand{\myparagraph}[1]{\vspace{0.1em}\noindent\textbf{#1}}
\usepackage{arydshln}
\newcommand \footnoteONLYtext[1]
{
	\let \mybackup \thefootnote
	\let \thefootnote \relax
	\footnotetext{#1}
	\let \thefootnote \mybackup
	\let \mybackup \imareallyundefinedcommand
}

\begin{document}
\title{Rethinking Boundary Detection in Deep Learning Models for Medical Image Segmentation}
\small
\author{Yi Lin$^{\dag}$, Dong Zhang$^{\dag}$, Xiao Fang, Yufan Chen, Kwang-Ting Cheng, and Hao Chen\textsuperscript{\Letter}}
\small
\institute{The Hong Kong University of Science and Technology, Hong Kong, China \\
\email{jhc@cse.ust.hk}
}
\titlerunning{Convolution \& Transformer \& Operator}
\authorrunning{Yi Lin, Dong Zhang, \emph{et al.}}
\maketitle 
\def\thefootnote{$\dag$}\footnotetext{Equal contribution; \Letter~corresponding author.}
\begin{abstract}
Medical image segmentation is a fundamental task in the community of medical image analysis. In this paper, a novel network architecture, referred to as Convolution, Transformer, and Operator (CTO), is proposed. CTO employs a combination of Convolutional Neural Networks (CNNs), Vision Transformer (ViT), and an explicit boundary detection operator to achieve high recognition accuracy while maintaining an optimal balance between accuracy and efficiency. The proposed CTO follows the standard encoder-decoder segmentation paradigm, where the encoder network incorporates a popular CNN backbone for capturing local semantic information, and a lightweight ViT assistant for integrating long-range dependencies. To enhance the learning capacity on boundary, a boundary-guided decoder network is proposed that uses a boundary mask obtained from a dedicated boundary detection operator as explicit supervision to guide the decoding learning process. The performance of the proposed method is evaluated on six challenging medical image segmentation datasets, demonstrating that CTO achieves state-of-the-art accuracy with a competitive model complexity.\footnoteONLYtext{The source code is available at \url{https://github.com/xiaofang007/CTO}}
\end{abstract}
\section{Introduction}
Medical Image Segmentation (MISeg) aims to locate pixel-level semantic lesion areas and/or human organs of the given image, which is one of the fundamental yet challenging tasks in the community of medical image analysis~\cite{ronneberger2015u,lin2021seg4reg}. 
In the past few years, this task has been extensively studied and applied to a wide range of downstream applications, \eg, robotic surgery~\cite{gao2021future}, cancer diagnosis~~\cite{lin2019automated}, and treatment design~\cite{wijeratne2021learning}. 
To achieve a desired MISeg result, it is critical to extract a set of rich and discriminative image feature representations. 

Recently, thanks to the successful utilization of Vision Transformer (ViT) on computer vision tasks~\cite{dosovitskiy2020image}, ViT-based methods have greatly promoted the accuracy of medical image analysis~\cite{hatamizadeh2022unetr}. For example, the state-of-the-art methods for some medical image analysis tasks (\eg, diagnosis~\cite{wu2022seatrans}, segmentation~\cite{chen2021transunet}, and detection~\cite{shamshad2022transformers}) are based on the ViT framework~\cite{dosovitskiy2020image}. Compared to CNNs-based methods, ViT has a stronger capacity to capture long-range dependencies, which have been shown to be beneficial for visual recognition~\cite{zhang2022graph}. For a canonical ViT-based MISeg model, it first partitions the input image into image patches. Then, these patches are treated as tokens for interactions via a multi-head self-attention layer, where the positional embedding is used for capturing the relative spatial information if needed. Finally, a normalization strategy and feature regulation operations are used to generate the output. The above processes are connected to form a basic transformer block, and such a block is repeated to encode semantic representations for the MISeg head network. 

Despite that ViT-based methods have achieved preliminary success, they inherently suffer from two potential problems, \ie, lack of translation invariance and weakness in local features~\cite{chen2021transunet}. To address these two problems, the advanced CNNs-ViT hybrid architectures were proposed for MISeg, \eg, TransUNet~\cite{chen2021transunet}, UNETR~\cite{hatamizadeh2022unetr}, Swin UNETR~\cite{hatamizadeh2022swin}. These attempts add convolutional operations in a ViT framework for local feature interactions, and strategies can improve the model convergence are also used. Particularly, the CNNs-ViT hybrid methods for MISeg are mainly based on UNet~\cite{ronneberger2015u} and add transformer blocks in the backbone networks~\cite{hatamizadeh2022unetr}, and skip connections~\cite{huang2021missformer}. 

The explicit boundary also matters - although this information is usually overlooked in the deep learning era. Compared to the implicit learning manner (\eg, CNNs, and ViT), an explicit learning model provides an immediate feature learning pattern, which has remarkable advantages of simple implementation, high efficiency, and purposeful objective. In the recent past, boundary operators are gradually valued in some pixel-level tasks and have been used to explicitly enhance the learning capacity on localization~\cite{chen2021image,fan2020camouflaged,lin2022label}. For MISeg, we believe that the boundary operator should play a more important role. Because, intuitively, a lesion region can be regarded as a kind of noise compared to normal regions. Besides, empirically, the explicit learning strategy can help the implicit feature learning model improve its representation capacity. 

We propose a new network architecture, called CTO (Convolution, Transformer, and Operator), forMISeg that combines CNNs, ViT, and boundary detection operators to leverage both local semantic information and long-range dependencies in the learning process. CTO follows the canonical encoder-decoder segmentation paradigm, where the encoder network is composed of a CNNs backbone and an assistant lightweight ViT branch. To enhance boundary learning capacity, we introduce a boundary-guided decoder network that uses a self-generated boundary mask extracted by boundary detection operators as explicit supervisions to guide the decoding learning process.
Our CTO architecture has higher recognition accuracy and achieves a better trade-off between accuracy and efficiency compared to the advanced MISeg architectures.
We evaluate CTO on six representative yet challenging MISeg datasets, \ie, two ISIC datasets~\cite{gutman2016skin,codella2019skin}, PH2~\cite{mendoncca2013ph}, CoNIC~\cite{graham2021conic}, LiTS17~\cite{bilic2019liver}, and BTCV~\cite{irshad2022improved}. Experimental results demonstrate that our CTO can achieve: 1) a new high accuracy on these datasets; 2) a superior performance to state-of-the-art methods; 3) and with competitive model complexity and efficiency. 
\section{Related Work}
\myparagraph{Medical Image Segmentation (MISeg).} The existing MISeg methods can be roughly divided into the following three camps:  \romannumeral1) CNNs-based methods; \romannumeral2) ViT-based methods; and \romannumeral3) CNNs-ViT hybrid methods. One of the most notable commonalities among these methods is that they are mainly based on an encoder-decoder paradigm. 
\textsf{In the first camp}, there are representative V-Net~\cite{milletari2016v}, U-Net~\cite{ronneberger2015u}, Attention-UNet~\cite{schlemper2019attention}. These methods use CNNs as the backbone to extract image features, and combine some elaborate tricks (\eg, skip connection, multi-scale representation~\cite{chen2016dcan}, feature interaction~\cite{chen2018voxresnet}) for feature enhancement. However, since convolution is inherently a local operation, methods in this camp may result in the problem of incomplete segmentation mask.
\textsf{In the second camp}, there are Swin-UNet~\cite{cao2021swin} and MissFormer~\cite{huang2021missformer}. Such methods use a ViT to replace CNNs as encoder/decoder to aggregate long-range feature dependencies. However, due to the limited number of medical images and the small inherent variability, such methods are difficult to optimize and have excessively high computational costs.
\textsf{In the third camp}, there are TransUNet~\cite{chen2021transunet}, UNETR~\cite{hatamizadeh2022unetr} and Swin UNETR~\cite{hatamizadeh2022swin}. This type of method combines advantages of both CNNs and ViT, \ie, the model can capture not only local information but also long-range feature dependencies. However, an obvious disadvantage is that they are computationally intensive and have high computation complexity. In our work, we propose to use a lightweight ViT as an assistant to help the mainstream CNN capture long-range feature dependencies. Besides, a boundary-enhanced feature, which is generated by an explicit boundary detection operator, is used to guide the decoding learning process.

\myparagraph{Operators in Image Processing.} The operator is a fundamental component in traditional digital image processing, where the boundary detection operator is the most core element. The commonly used boundary detection operators can be divided into: \romannumeral1) the first derivative operator (\eg, Roberts, Prewitt, and Sobel), and \romannumeral2) the second derivative operators (\eg, Laplacian)~\cite{kanopoulos1988design}. 
Recently, boundary detection operators have been revived in pixel-level computer vision tasks, such as manipulation detection~\cite{chen2021image} and camouflaged object detection~\cite{fan2020camouflaged}.
In this paper, the boundary detection operator is used as an explicit mask extractor to guide an implicit feature learning model for MISeg. Our contribution is to use feature maps of the intermediate layer to synthesize a high-quality boundary prediction without requiring additional information. 
\section{Convolution, Transformer, and Operator (CTO)}
\label{sec:method}
\subsection{Overview}
\label{sec:arch}
\begin{figure}[t]
\centering
\includegraphics[width=.98\textwidth]{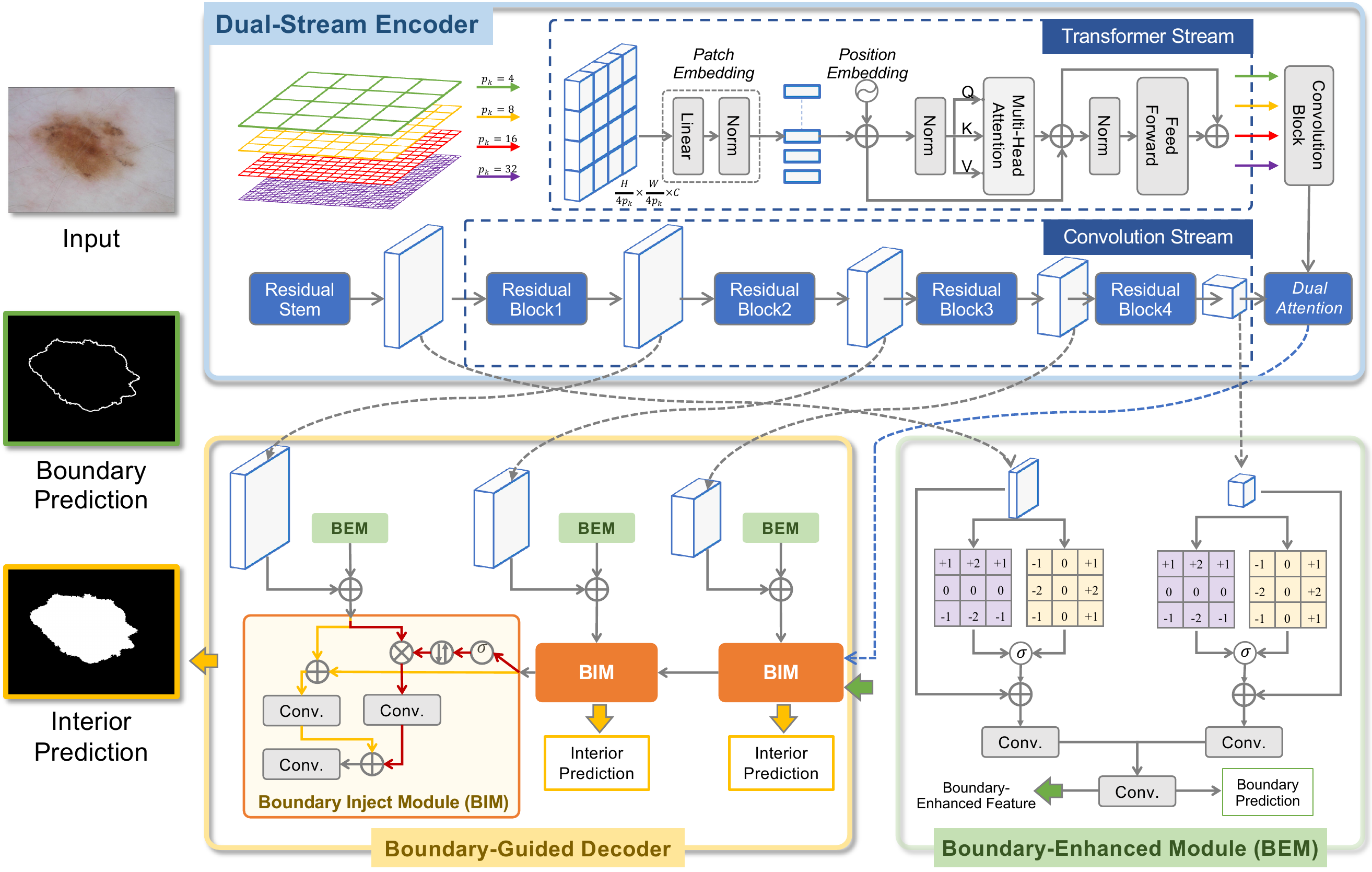}
\caption{Illustration of our CTO, which follows an encoder-decoder paradigm, where the \textsf{ encoder network} consists of a mainstream CNNs and an assistant ViT. The \textsf{decoder network} employs a boundary detection operator to guide its learning process.}
\label{fig:framework}
\end{figure}
The overall architecture of CTO is illustrated in Figure~\ref{fig:framework}. For an input image $X \in \mathbb{R}^{H\times W\times 3}$ with a spatial resolution of $H\times W$ and $C$ channels, we aim to predict a pixel-wise labelmap $Y$, where each pixel has been assigned a class label. The whole model follows an encoder-decoder pattern, which also adopts skip connections to aggregate low-level features from the encoder to the decoder. For the encoder, we design a dual-stream encoder (\textit{ref.}~Sec.~\ref{sec:encoder}), which combines a convolutional neural network (\ie, Res2Net~\cite{gao2019res2net}) and a lightweight vision transformer to capture local feature dependencies and long-range feature dependencies between image patches, respectively. Such a combination will not bring many computational overheads. For the decoder, an operator-guided decoder (\textit{ref.}~Sec.~\ref{sec:decoder}) uses a boundary detection operator (\ie, Sobel~\cite{kanopoulos1988design}) to guide the learning process via the generated boundary mask. The whole model is trained in an end-to-end manner.
\subsection{Dual-Stream Encoder}
\label{sec:encoder}
\subsubsection{The Mainstream Convolution Stream.}
The convolution stream is used to capture local feature dependencies. To this end, we choose the strong yet efficient Res2Net~\cite{gao2019res2net} as the backbone, which is composed of one convolution stem and four residual blocks, generating feature maps $F_c^k$ with the spatial resolution of $H/4\times W/4$, $H/8\times W/8$, $H/16\times W/16$, and $H/32\times W/32$, respectively. 

\subsubsection{The Assistant Transformer Stream.}
\label{sec:mst}
The lightweight vision transformer (LightViT) is designed to capture the long-range feature dependencies between image patches in different scales.
Specifically, the LightViT consists of multiple parallel lightweight transformer blocks that are fed with feature patches in different scales.
All the transformer blocks share a similar structure, which consists of patch embedding layers and transformer encoding layers.

As shown in Figure~\ref{fig:framework}, given the input feature map $\text{F}_1^c \in \mathbb{R}^{\frac{H}{4}\times \frac{W}{4}\times C}$, we first divide it into $\frac{HW}{16p^2}$ patches with size $p\times p$, and then flatten each patch into a vector $\mathbf{v}_i \in \mathbb{R}^{p^2\times C}$.
In our paper, we use four parallel transformer blocks, which are fed with feature patches in size of $p=4,8,16,32$.
Then, we apply a linear projection to each patch vector to obtain the patch embedding $\text{e}_i \in \mathbb{R}^{C}$.
After that, patch embeddings along with the position embeddings are fed into the transformer encoding layers to obtain the output. Following~\cite{dosovitskiy2020image}, the encoding layers consist of a lightweight multi-head self-attention (MHSA) layer and a feed-forward network.
MHSA receives a truncated query $Q$, key $K$, and value $V$ as input, and then computes the attention score $\text{A} \in \mathbb{R}^{N\times N}$ as follows:
\begin{equation}
    \text{A} = \text{softmax}\left(\frac{QK^T}{\sqrt{d_k}}\right)V,
\end{equation}
where $N$ is the size of patch number, $d_k$ is the dimension of the key.
The output of the MHSA layer is then fed into a feed-forward layer to obtain output $\text{F}_t$:
\begin{equation}
    \text{F}_t = \text{FFN}(\text{A}),
\end{equation}
where $\text{FFN}$ is the feed-forward network with two linear layers with ReLU activation function.
Then, $\text{F}_t$ is reshaped into the same size as $\text{F}_c^1$ to obtain the output. 
Outputs of all the transformer blocks are concatenated along the channel dimension and fed into the convolutional layer to obtain the final output.
\subsection{Boundary-Guided Decoder}
\label{sec:decoder}
The boundary-guided decoder uses a gradient operator module to extract the boundary information of foreground objects. 
Then, the boundary-enhanced feature $F_b$ is integrated into multi-level encoder's features by a boundary optimization module, aiming to simultaneously characterize the intra- and inter-class consistency in the feature space, enriching the feature representative ability.
\subsubsection{Boundary Enhanced Module (BEM).}
BEM takes the high-level $F_c^4$ and low-level features $F_c^1$ as inputs to extract the boundary information while filtering the trivial boundary irrelevant information. To achieve this goal, we apply Sobel operator~\cite{kanopoulos1988design} at both horizontal $G_x$ and vertical $G_y$ directions to obtain the gradient maps. Specifically, we utilize two $3\times 3$ parameter-fixed convolutions and apply convolution operation with stride 1. Two convolutions are defined as:
\begin{equation}
    K_x = \begin{bmatrix}
        -1 & 0 & 1 \\
        -2 & 0 & 2 \\
        -1 & 0 & 1
    \end{bmatrix}, \quad
    K_y = \begin{bmatrix}
        -1 & -2 & -1 \\
        0 & 0 & 0 \\
        1 & 2 & 1
    \end{bmatrix}.
\end{equation}

Then, we apply the two convolutions to the input feature map to obtain the gradient maps $M_x$ and $M_y$.
After that, the gradient maps are normalized by a sigmoid function and then fused with the input feature map to obtain the edge-enhanced feature map $F_e$:
\begin{equation}
    F_e = F_c \odot \sigma(M_{xy}),
\end{equation}
where $\odot$ denotes the element-wise multiplication, $\sigma$ is the sigmoid function, and $M_{xy}$ is the concatenation of $M_x$ and $M_y$ along the channel dimension.
Then, we fuse the edge-enhanced feature maps of $F_e^1$ and $F_e^4$ with a simple stacked convolution layer in the bottleneck.
Specifically, we first apply a $1\times 1$ convolution with 
a bilinear upsampling operation to the feature map $F_e^4$ to obtain the feature map with the same size as $F_e^1$.
Then, we separately apply $1\times 1$ convolution operation to equate the channel size of these two features.
Finally, we concatenate these two feature maps along the channel dimension and apply a two-layer convolutions to get the final feature map $\bar{F}_e$.
The output is supervised by the ground truth boundary map, which in turn eliminates the edge feature inside the objects, producing the boundary-enhanced feature $F_b$.


\subsubsection{Boundary Inject Module (BIM).}
The obtained boundary-enhanced feature from BEM can be used as a prior to improve the image representation ability of the features produced by the encoder.
We propose BIM that introduces a dual path boundary fusion scheme to promote the feature representation in both foreground and background.
Specifically, BIM takes two inputs: the channel-wise concatenation of the boundary-enhanced feature $F_b$ and the corresponding feature $F_c$ from the encoder network, and the feature from the previous decoder layer $F_d^{j-1}$.
Then, these two inputs are fed into BIM, which contains two individual paths aiming to promote the feature representation in the foreground and background, respectively.
For the foreground path, we directly concatenate the two inputs along the channel dimension, and then apply a sequential \textsf{Conv-BN-ReLU} (\ie, convolution, batch normalization, ReLU activator) layers to obtain the foreground feature $F_{fg}$.
For the background path, we design the background attention component to selectively focus on the background information, which is expressed as:
\begin{equation}
    F_{bg} = \text{Convs}\left ((1-\sigma (F_d^{j-1})) \odot F_c \right ),
\end{equation}
where $\text{Convs}$ is a three-layer \textsf{Conv-BN-ReLU} layers, $\sigma$ is the sigmoid function, and $\odot$ denotes the element-wise multiplication.
The term $\left(1-(\sigma F_d^{j-1})\right)$ is the background attention map, which is computed by first applying the sigmoid function to the feature map from the previous decoder layer, which will generate a foreground attention map.
Then, we subtract the foreground attention map from 1 to obtain the background attention map.
Finally, we concatenate the foreground feature $F_{fg}$, the background feature $F_{bg}$, and the previous decoder feature $F_d^{j-1}$ along the channel dimension to obtain the final output $F_d^j$.
\subsection{Overall Loss Function}
\label{sec:allloss}
\label{sec:loss}
Since the proposed CTO is a multi-task model (\ie, interior and boundary segmentation), we define an overall loss function to jointly optimize these two tasks.

\subsubsection{Interior Segmentation Loss.}
The interior segmentation loss is the weighted sum of cross-entropy loss $\mathcal{L}_{\text{CE}}$ and mean intersection-over-union (mIoU) loss $\mathcal{L}_{\text{mIoU}}$, which are defined as:
\begin{equation}
    \mathcal{L}_{\text{CE}} = -\frac{1}{N}\sum_{i=1}^N \left(y_i \log(\hat{y}_i) + (1-y_i) \log(1-\hat{y}_i) \right),
\end{equation}
\begin{equation}
    \mathcal{L}_{\text{mIoU}} = 1 - \frac{\sum_{i=1}^N (y_i * \hat{y}_i)}{\sum_{i=1}^N (y_i + \hat{y}_i - y_i * \hat{y}_i)},
\end{equation}
where $y_i$ and $\hat{y}_i$ are the ground truth and the predicted label for the $i$-th pixel, respectively, and $N$ is the total number of pixels of the image.

\subsubsection{Boundary Loss.}
Considering the class imbalance problem between the foreground and background pixels in boundary detection, we employ the Dice Loss:
\begin{equation}
    \mathcal{L}_{\text{Dice}} = 1 - \frac{2 \sum_{i=1}^N (y_i * \hat{y}_i)}{\sum_{i=1}^N (y_i + \hat{y}_i)}.
\end{equation}

\subsubsection{Total Loss.}
The total loss is composed of the major segmentation loss $\mathcal{L}_{\text{seg}}$, and the boundary loss $\mathcal{L}_{\text{bnd}}$. 
Note that for the boundary detection loss, we only consider the prediction from BEM, which takes encoder's feature maps from the high-level layer (\ie, $F_b^4$) and low-level layer (\ie, $F_b^1$) as input.
As for the major image segmentation loss, we apply the deep supervision strategy to obtain the prediction from the decoder's feature at different levels.
In summary, the total loss can be formulated as:
\begin{equation}
    \mathcal{L} = \mathcal{L}_{\text{seg}} + \mathcal{L}_{\text{bnd}}
                = \sum_i^L \left(\mathcal{L}_{\text{CE}} + \mathcal{L}_{\text{mIoU}}\right)  + \alpha\mathcal{L}_{\text{Dice}},
\end{equation}
where $L$ is the number of BOMs, which is set to 3 in this work. $\alpha$ is the weighting factor, which is set to 3 to balance the losses.

\section{Experiments}
\label{sec:exp}
\subsection{Datasets and Evaluation Metrics}
\label{sec:dataset}
\myparagraph{Datasets.} We evaluate our CTO on six public MISeg datasets, including three datasets for skin lesion segmentation, \ie, ISIC~\cite{gutman2016skin,codella2019skin} and PH2~\cite{mendoncca2013ph}, the Colon Nuclei Identification and Counting (CoNIC) challenge dataset~\cite{graham2021conic}, the Liver Tumor Segmentation (LiTS17) Challenge dataset~\cite{bilic2019liver}, and the Beyond the Cranial Vault (BTCV) challenge dataset~\cite{vaswani2017attention}. 
As in~\cite{gutman2016skin,codella2019skin}, we perform 5-fold cross-validation on ISIC 2018, and train the model on ISIC 2016 and test it on PH2~\cite{mendoncca2013ph}. BTCV is divided into 18 cases for training and 12 cases for test~\cite{cao2021swin,chen2021transunet}. CoNIC and LiTS17 are randomly divided into training, validation, and test sets with a radio of 7:1:2.
\myparagraph{Evaluation Metrics.} Following~\cite{cao2021swin,chen2021transunet,lee2020structure}, the commonly used  Dice Coefficient (Dice), Intersection over Union (IoU), average Hausdorff Distance (HD) and Panoptic Quality (PQ) are used as the primary accuracy evaluation metrics. Besides, FLOPs and model parameters are used to evaluate the model efficiency. 
\subsection{Implementation Details}
\label{sec:implementation}
We optimize our model using the ADAM optimizer with an initial learning rate 1e-4.  The default batch size is set to 32 with the image size of 256$\times$256.
The encoder is initialized with the pre-trained weights of Res2Net-50~\cite{gao2019res2net} on ImageNet and then fine-tuned for 90 epochs on a single NVIDIA RTX 3090 GPU. 
All 3D volumes are inferenced in a sliding-window manner with the stride of 1, and the final segmentation results are obtained by stacking the prediction maps to reconstruct the 3D volume for evaluation.
Except for a special statement, all the experimental settings follow the baseline paper~\cite{cao2021swin,chen2021transunet,lee2020structure,zhang2022deep}.
\subsection{Experimental Results}
\label{sec:results}
\begin{figure}[t]
\centering
\includegraphics[width=.95\textwidth]{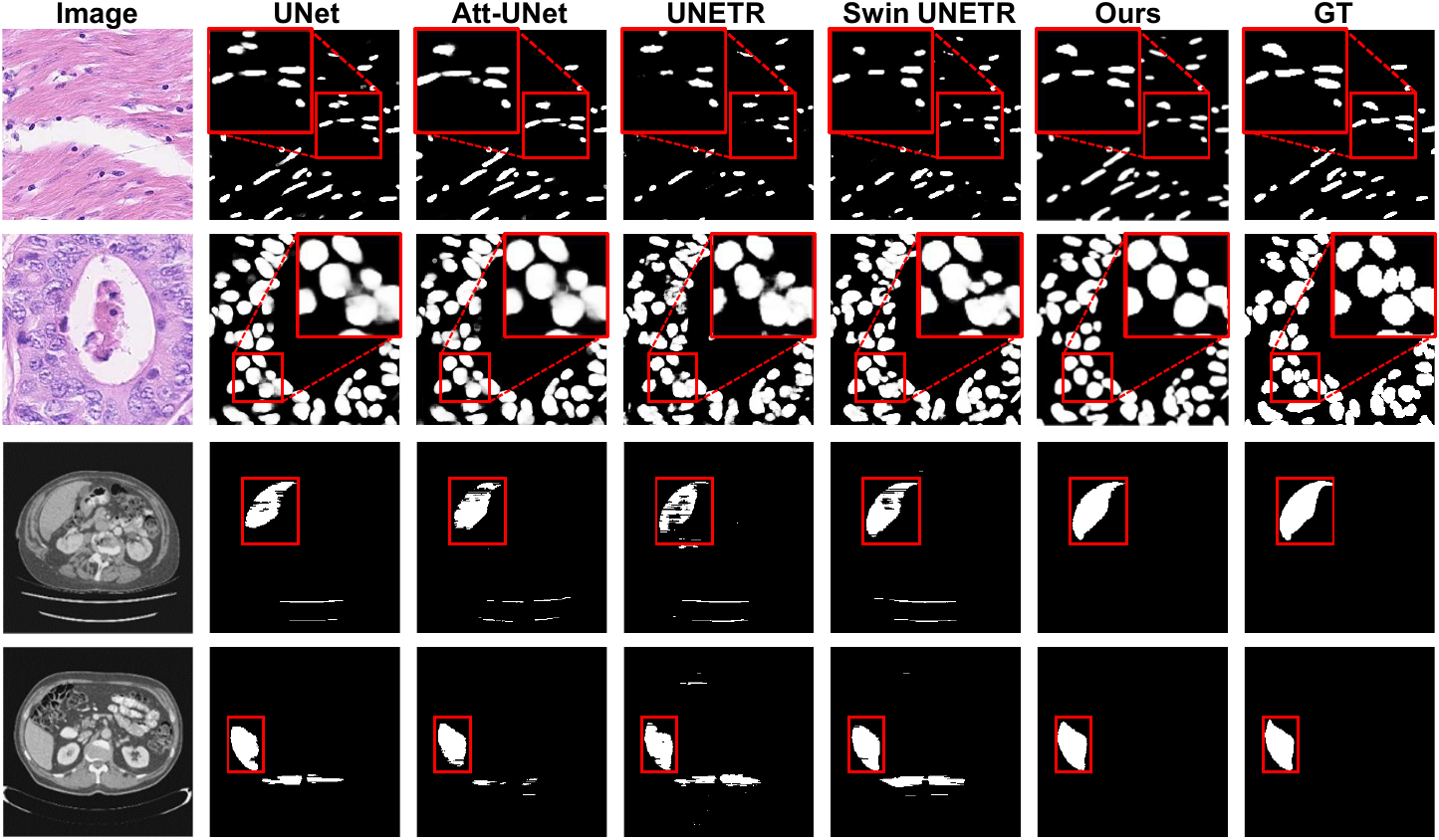}
\caption{Visualizations on CoNIC~\cite{graham2021conic} (top two rows) and LiTS17~\cite{bilic2019liver} (buttom two rows). The \textcolor{red}{red} boxes highlight the main difference of each method.}
\label{fig:visual_conic}
\end{figure}
\begin{table}[t]
\caption{Comparisons with other methods on ISIC~\cite{gutman2016skin,codella2019skin} \& PH2~\cite{mendoncca2013ph}.}
\label{tab:sota_isic}
\centering
\renewcommand\arraystretch{1.2}
\setlength{\tabcolsep}{4pt}{
\begin{tabular}{p{2cm}<{\centering}p{1.5cm}<{\centering}p{1.5cm}<{\centering}|p{2cm}<{\centering}p{1.5cm}<{\centering}p{1.5cm}<{\centering}}
\hline
\hline
\multirow{2}{*}{Methods} & \multicolumn{2}{c|}{ISIC 2016 \& PH2} & \multirow{2}{*}{Methods} & \multicolumn{2}{c}{ISIC 2018}\\
 & Dice~$\uparrow$ & IoU~$\uparrow$ &  & Dice~$\uparrow$ & IoU~$\uparrow$ \\
\hline
SSLS\cite{ahn2015automated} & 78.38 & 68.16 & Deeplabv3~\cite{chen2017rethinking}    & 88.4& 80.6\\
MSCA\cite{bi2016automated} & 81.57 & 72.33 & U-Net++~\cite{zhou2018unetpp}          & 87.9& 80.5\\
FCN~\cite{long2015fully} & 89.40 & 82.15 & CE-Net~\cite{gu2019cenet}              & 89.1& 81.6\\
Bi~\textit{et al}~\cite{bi2017dermoscopic} & 90.66 & 83.99 & MedT~\cite{valanarasu2021medical}      & 85.9 & 77.8\\
Lee~\textit{et al}~\cite{lee2020structure} & \underline{91.84} & \underline{84.30} &TransUNet~\cite{chen2021transunet}      & \underline{89.4}& \underline{82.2}\\
\cdashline{1-6}[1.5pt/1.5pt]
CTO(Ours) & \textbf{91.89} & \textbf{85.18} & Ours & \textbf{91.2} & \textbf{84.5} \\
\hline
\hline
\end{tabular}}
\end{table}
\begin{table}[t]
\caption{Comparisons with other methods on CoNIC~\cite{graham2021conic} and LiTS17~\cite{bilic2019liver}.}
\label{tab:sota_conic_lits}
\centering
\renewcommand\arraystretch{1.2}
\setlength{\tabcolsep}{3pt}{
\begin{tabular}{p{2.5cm}|p{1cm}<{\centering}p{1cm}<{\centering}p{1cm}<{\centering}|p{1cm}<{\centering}p{1cm}<{\centering}|p{1.5cm}<{\centering}p{1.5cm}<{\centering}}
\hline
\hline
\multirow{2}{*}{Methods} & \multicolumn{3}{c|}{CoNIC} & \multicolumn{2}{c|}{LiTS17} & \multicolumn{2}{c}{Model Efficiency}\\
 & Dice~$\uparrow$ & IoU~$\uparrow$ & PQ~$\uparrow$  & Dice~$\uparrow$ & IoU~$\uparrow$ & \scriptsize{Param.(M)} & \scriptsize{GFLOPs}\\
\hline
V-Net~\cite{milletari2016v}                             & 77.46 & 64.94 & 63.59 & 89.20  & 80.71 & 11.84 & 18.54\\
U-Net~\cite{ronneberger2015u}                           & 78.42 & \underline{66.39} & 64.44 & 84.66  & 73.63 & 7.78  & 14.59\\
R50-UNet~\cite{ronneberger2015u}                        & 77.67 & 65.34 & 63.67 & \underline{91.24}  & \underline{84.14} & 33.69 & 20.87\\
Att-UNet~\cite{schlemper2019attention}                  & \underline{79.48} & 66.06 & \underline{65.25} & 85.88  & 75.40 & 7.88  & 43.35\\
R50-\scriptsize{AttUNet}~\cite{schlemper2019attention}  & 78.21 & 65.86 & 64.02 & 89.98  & 82.13 & 33.25 & 49.25\\
R50-ViT~\cite{dosovitskiy2020image}                     & 75.36 & 62.35 & 58.03 & 83.67  & 72.49 & 110.62 & 26.91\\
UNETR~\cite{hatamizadeh2022unetr}                      & 71.46 & 57.24 & 52.26 & 81.48  & 69.04 & 87.51 & 26.41\\
\scriptsize{Swin-UNETR}~\cite{hatamizadeh2022swin}      & 70.07 & 55.56 & 51.59 & 84.00  & 72.76 & 6.29  & 4.86\\
\cdashline{1-8}[1.5pt/1.5pt]
CTO(Ours)                                               & \textbf{79.77} & \textbf{66.42} & \textbf{65.58} & \textbf{91.50} & \textbf{84.59} & 59.82 & 22.72\\
\hline
\hline
\end{tabular}}
\end{table}
\begin{table}[t]
\caption{Comparisons with other methods on BTCV~\cite{irshad2022improved}.}
\footnotesize
\renewcommand\arraystretch{1.2}
\setlength{\tabcolsep}{.1pt}{
\begin{tabular}{l|m{1cm}<{\centering}m{1cm}<{\centering}|m{1cm}m{1cm}m{1cm}m{1cm}m{1cm}m{1cm}m{1cm}m{1cm}}
\hline
\hline
Methods & mDice$\uparrow$ & HD$\downarrow$ & Aorta & Gallb. &\scriptsize{Kid(L)} & \scriptsize{Kid(R)} & Liver & Panc. & Spleen & Stom. \\
\hline
V-Net~\cite{milletari2016v}     & 68.81        & -            & 75.34          & 51.87              & 77.10             & \underline{80.75}             & 87.84           & 40.05            & 80.56          & 56.98           \\
DARR~\cite{fu2020domain}      & 69.77        & -            & 74.74           & 53.77             & 72.31            & 73.24             & 94.08           & 54.18            & 89.90           & 45.96            \\
U-Net~\cite{ronneberger2015u}   & 76.85 & 39.70 & \underline{89.07} & \textbf{69.72} & 77.77 & 68.60 & 93.43 & 53.98 & 86.67 & 75.58 \\
R50-UNet~\cite{ronneberger2015u}             & 74.68        & 36.87        & 84.18      & 62.84          & 79.19         & 71.29          & 93.35        & 48.23       & 84.41       & 73.92        \\
Att-UNet~\cite{schlemper2019attention} & 77.77 & 36.02 & \textbf{89.55} & \underline{68.88} & 77.98 & 71.11 & 93.57 & \underline{58.04} & 87.30 & 75.75 \\
R50-\scriptsize{AttUNet}~\cite{schlemper2019attention} & 75.57   & 36.97        & 55.92      & 63.91          & 79.20         & 72.71          & 93.56       & 49.37        & 87.19       & 74.95        \\
R50-ViT~\cite{dosovitskiy2020image}    & 71.29        & 32.87        & 73.73      & 55.13          & 75.80         & 72.20         & 91.51       & 45.99        & 81.99      & 73.95        \\
TransUNet~\cite{chen2021transunet} & 77.48        & 31.69        & 87.23        & 63.13          & 81.87          & 77.02          & 94.08      & 55.86         & 85.08      & 75.62        \\ 
SwinUNet~\cite{cao2021swin} & \underline{79.12} & \underline{21.55} & 85.47 & 66.53 & \underline{83.28} & 79.61 & \underline{94.29} & 56.58 & \textbf{90.66} & \underline{76.60} \\
\cdashline{1-11}[1.5pt/1.5pt]
CTO(Ours) & \textbf{81.10} & \textbf{18.75} & 87.72 & 66.44 & \textbf{84.49} & \textbf{81.77} & \textbf{94.88} & \textbf{62.74} & \underline{90.60} & \textbf{80.20} \\
\hline
\hline
\end{tabular}
\vspace{-0.6em}
}
\label{tab:sota_btcv}
\end{table}

\subsubsection{Comparisons with State-of-the-Art Methods.}
We compare our CTO with the state-of-the-art (SOTA) methods including U-Net~\cite{ronneberger2015u}, ResUNet~\cite{ronneberger2015u} with ResNet-50~\cite{he2016deep} as the backbone, VNet~\cite{milletari2016v}, ViT~\cite{dosovitskiy2020image}, TransUNet~\cite{chen2021transunet}, and Swin-Unet~\cite{cao2021swin}. 
On ISIC 2016~\cite{gutman2016skin} \& PH2~\cite{mendoncca2013ph}, we compare CTO with five related methods.
The results are shown in Table~\ref{tab:sota_isic}.
We can observe that CTO achieves 91.89\% in Dice and 85.18\% in IoU, which outperforms the SOTA methods by 0.05\% and 0.88\%, respectively.
On ISIC 2018~\cite{codella2019skin}, our CTO achieves 91.2\% in Dice and 84.5\% in IoU by 5-fold cross-validation, which outperforms the SOTA methods by 1.8\% and 2.3\%, respectively.
On CoNIC~\cite{graham2021conic}, 
results are shown in Table~\ref{tab:sota_conic_lits}, we can observe that our CTO achieves 79.77\%, 66.42\%, and 65.58\% in Dice, IoU, and PQ, respectively, consistently outperforming other methods.
Qualitative result comparisons are illustrated in Figure~\ref{fig:visual_conic}. We can observe that our CTO delineates more accurate object contours than other methods regarding diverse shapes and sizes of nuclei, especially on some blurred nuclei objects.

We also conduct experiments on 3D MISeg tasks.
On LiTS17~\cite{bilic2019liver}, as shown in Table~\ref{tab:sota_conic_lits}, our model achieves 91.50\% in Dice and 84.59\% in IoU, outperforming SOTA methods by 0.26\% and 0.45\%, respectively.
On BTCV~\cite{irshad2022improved}, as shown in Table~\ref{tab:sota_btcv}, our CTO achieves 81.10\% in Dice and 18.75\% in HD, which outperforms the SOTA methods.
In particular, as for Dice, our CTO outperforms the second, third, and fourth best methods by 1.98\%, 3.33\%, and 3.62, respectively.
Besides, the distinct improvements can be markedly observed for organs with blurry boundaries, \eg, the ``pancreas'' and the ``stomach'', where our model achieves significant gains over the SOTA methods, \ie, 4.70\% and 3.60\% in Dice, respectively.
As for the model efficiency, we can observe that CTO achieves competitive  performance improvements with comparable FLOPs and parameters. 
\begin{table}[t]
    \caption{Ablation study results on ISIC 2018~\cite{codella2019skin}. * means the component achieves significant performance improvement with p < 0.05 via paired t-test.}
    \vspace{-2mm}
    \label{tab:abl_isic}
    \centering
    \renewcommand\arraystretch{1}
    \setlength{\tabcolsep}{5pt}{
    \renewcommand\arraystretch{1.2}
    \begin{tabular}{p{1cm}<{\centering}p{1cm}<{\centering}p{1cm}<{\centering}p{1cm}<{\centering}p{1.5cm}<{\centering}|p{1.5cm}<{\centering}p{2cm}<{\centering}}
    \hline
    \hline
    CNNs & LightViT & CBM & BEM & BIM & Dice~$\uparrow$ & IoU~$\uparrow$ \\
    \hline
    \cmark  &           &           &           &           & 88.32             & 81.51 \\  
    \cmark  & \cmark    &           &           &           & 89.31$^*_{\color{red}{+0.99}}$    & 82.47$^*_{\color{red}{+0.96}}$ \\
    \cmark  & \cmark    & \cmark    &           &           & 89.41$^*_{\color{red}{+1.09}}$    & 82.51$_{\color{red}{+1.00}}$ \\
    \cmark  & \cmark    & \cmark    & \cmark    &           & 89.52$_{\color{red}{+1.20}}$    & 82.81$_{\color{red}{+1.30}}$ \\
    \cmark  & \cmark    & \cmark    & \cmark    & \cmark    & 91.21$^*_{\color{red}{+2.89}}$    & 84.45$^*_{\color{red}{+2.94}}$ \\
    \hline
    \hline
    \end{tabular}}
    \vspace{-3mm}
\end{table}
\myparagraph{Ablation Study.}
We conduct ablation studies to explore the effectiveness of each component in CTO. In Table~\ref{tab:abl_isic}, we compare the performance of CTO variants on ISIC 2018~\cite{codella2019skin}: 1) CNNs, only the convolution stream; 2) $+$LightViT, the dual-stream encoder with convolution and Transformer; 3) $+$CBM, adding the boundary supervision with the same architecture of BEM, except the Sobel layer; 4) $+$BEM, the boundary-enhanced module; 5) $+$BIM, the boundary inject module. 
All the components consistently boost the performance by 0.99\%, 1.09\%, 1.20\%, 2.89\% in Dice, respectively.
Especially, we observe that the boundary supervision (\ie, BIM) is crucial for MISeg.

\section{Conclusion}
In this study, a new network architecture named CTO is proposed for MISeg. Compared to advanced MISeg architectures, CTO achieves a better balance between recognition accuracy and computational efficiency. The contribution of this paper is the utilization of intermediate feature maps to synthesize a high-quality boundary supervision mask without requiring additional information. Results from experiments conducted on six publicly available datasets demonstrate the superiority of CTO over state-of-the-art methods, and the effectiveness of each of its components. Future work includes the extension of the concept of a couple-stream encoder to various advanced backbone architectures, and the potential adaptation of CTO to a 3D manner.

\bibliographystyle{splncs04}
\bibliography{ref_pub.bib}
\end{document}